%% file: main.tex
\documentclass{article}

\usepackage{microtype}
\usepackage{graphicx}
\usepackage{subfigure}
\usepackage{booktabs} 

\usepackage{hyperref}
\usepackage{multirow}

\usepackage[accepted]{icml2025}

\usepackage{amsmath}
\usepackage{amssymb}
\usepackage{mathtools}
\usepackage{amsthm}
\usepackage{subcaption}
\usepackage[capitalize,noabbrev]{cleveref}

\usepackage{arydshln}

\theoremstyle{plain}

\theoremstyle{definition}

\theoremstyle{remark}

\usepackage[textsize=tiny]{todonotes}

\icmltitlerunning{Unlocking LLM Potential via Aggregation Fine-Tuning}
\begin{document}

\twocolumn[
\icmltitle{From Drafts to Answers: \\Unlocking LLM Potential via Aggregation Fine-Tuning}



\icmlsetsymbol{equal}{*}

\begin{icmlauthorlist}
\icmlauthor{Yafu Li}{equal,shai}
\icmlauthor{Zhilin Wang}{equal,shai}
\icmlauthor{Tingchen Fu}{rcu}
\icmlauthor{Ganqu Cui}{shai}
\icmlauthor{Sen Yang}{cuhk}
\icmlauthor{Yu Cheng}{cuhk}
\end{icmlauthorlist}

\icmlaffiliation{shai}{Shanghai AI Laboratory}
\icmlaffiliation{rcu}{Renmin University of China}
\icmlaffiliation{cuhk}{The Chinese University of Hong Kong}

\icmlcorrespondingauthor{Yu Cheng}{chengyu@cse.cuhk.edu.hk}

\icmlkeywords{Machine Learning, ICML}

\vskip 0.3in
]



\printAffiliationsAndNotice{\icmlEqualContribution} 

\begin{abstract}
Scaling data and model size has been proven effective for boosting the performance of large language models. 
In addition to training-time scaling, recent studies have revealed that increasing test-time computational resources can further improve performance.
In this work, we introduce \emph{Aggregation Fine-Tuning} (AFT), a supervised fine-tuning paradigm where the model learns to synthesize multiple draft responses, referred to as \emph{proposals}, into a single, refined answer, termed \emph{aggregation}. 
At inference time, a \emph{propose-and-aggregate} strategy further boosts performance by iteratively generating proposals and aggregating them. 
Empirical evaluations on benchmark datasets show that AFT-trained models substantially outperform standard SFT. 
Notably, an AFT model, fine-tuned from \textbf{Llama3.1-8B-Base} with only \textbf{64k data}, achieves a 41.3\% LC win rate on AlpacaEval 2, surpassing significantly larger LLMs such as \textbf{Llama3.1-405B-Instruct} and GPT-4.
By combining sequential refinement and parallel sampling, the propose-and-aggregate framework scales inference-time computation in a flexible manner. 
Overall, These findings position AFT as a promising approach to unlocking additional capabilities of LLMs without resorting to increasing data volume or model size. 
Our code is publicly available at \href{https://github.com/Linzwcs/AFT}{https://github.com/Linzwcs/AFT}.
\end{abstract}


\input{tex_sections/1-Introduction}

\input{tex_sections/2-RelatedWork}
\input{tex_sections/3-Method}
\input{tex_sections/3.5-Experimental-Setup}

\input{tex_sections/4-Experiments}

\input{tex_sections/5-Analysis}
\input{tex_sections/6-Conclusion}


\section*{Impact Statement}
This work aims to advance the field of Machine Learning by providing a new training and inference paradigm, i.e., Aggregation Fine-Tuning, that refines multiple candidate answers into a stronger overall solution.
While we do not foresee any immediate or significant societal harms unique to this approach, it is possible that more powerful language generation techniques could be misused for deceptive or harmful purposes (e.g., generating misleading content). 
We encourage future efforts to incorporate robust filtering and responsible usage guidelines for such systems. 
Beyond these considerations, the potential positive societal impacts include more accurate, reliable, and resource-efficient language models, which could benefit a wide range of applications from education to healthcare.





\bibliography{main}
\bibliographystyle{icml2025}

\newpage
\appendix
\onecolumn
\section{Prompt}
\label{app:prompt}
We use the same aggregation prompt as MoA~\cite{moa} to perform aggregation, as shown in Table~\ref{tab:prompt}.
The proposals are embedded within this template and input into LLMs as a system prompt for aggregation.

\begin{table*}[t]
\caption{A specialized prompt to construct aggregation responses given multiple proposals.}
\label{tab:prompt}
\vskip 0.15in
\begin{center}
\begin{small}
\begin{tabular}{@{}p{1\linewidth}@{}}
    \toprule
    You have been provided with a set of responses from various distributions to the latest user query. Your task is to synthesize these responses into a single, high-quality response. It is crucial to critically evaluate the information provided in these responses, recognizing that some of it may be biased or incorrect. Your response should not simply replicate the given answers but should offer a refined, accurate, and comprehensive reply to the instruction. Ensure your response is well-structured, coherent, and adheres to the highest standards of accuracy and reliability.
    \\ \\
    Responses from models:
    \\
    1. [Model Response from $r_{i}$] \\
    2. [Model Response from $r_{i}$] \\
    ... \\
    $n$. [Model Response from $r_{n}$] \\
    \\
    User query: [query] \\
    \bottomrule
\end{tabular}
\end{small}
\end{center}
\vskip -0.1in
\end{table*}

    %



\section{Case Study}
\label{app:case}

We present case studies that illustrate the enhancement of generation quality through propose-and-aggregate across various tasks, including mathematics (Table~\ref{tab:case_math}), reasoning (Table~\ref{tab:case_reasonning}), knowledge (Table~\ref{tab:case_knowledge}), writing (Table~\ref{tab:case_writing}), and role-play (Table~\ref{tab:case_role_play}).

\section{Computational Overhead}
\label{app:flops}

Following~\citet{llm_monkey}, the FLOPs for generation can be approximated as:
\[
\text{FLOPs per token} \approx 2 \times (\text{num parameters} + 2 \times \text{num layers} \times \text{token dim} \times \text{context length}),
\]
and the total inference FLOPs are estimated as:
\[
\text{total inference FLOPs} \approx (\text{num prompt tokens} + \text{num decoded tokens}) \times \text{FLOPs per token}.
\]

For simplicity and without loss of generality, we omit the FLOPs for the query prompt tokens and focus on the computational overhead contributed by prompt tokens accommodating previous-layer proposals. Under this assumption:

For vanilla generation, the total inference FLOPs can be approximated as:
\[
\text{total inference FLOPs} \approx \text{num decoded tokens} \times \text{FLOPs per token}.
\]

For parallel sampling with a reward model, the FLOPs include the cost of generating multiple proposals in parallel and evaluating them using the RM. The total FLOPs, denoted as \(\bar{F}\), can be approximated as:
\[
\bar{F} \approx 2 \cdot (\text{num proposals}) \cdot (\text{num decoded tokens}) \cdot \text{FLOPs per token}.
\]
Here, the factor of 2 accounts for the additional computational cost of evaluating the proposals using an RM, which is typically comparable in size to the policy model.

For the propose-and-aggregate framework, the additional overhead comes from the aggregation step, where all proposals from the previous layer are included as part of the input prompt. Considering
the practical implementations of KV Cache that enable keys and values of contextual tokens (i.e., proposals from the previous layer) to be reusable, the total inference FLOPs, denoted as \(\hat{F}\), can be approximated as:
\[
\hat{F} \approx (\text{num aggregation layers}) \cdot (2 \cdot \text{num proposals} \cdot \text{num decoded tokens} \cdot \text{FLOPs per token})
\]
\[
+ (\text{num decoded tokens} \cdot \text{FLOPs per token}),
\]
 where \((2 \cdot \text{num aggregation layers} \cdot \text{num proposals} \cdot \text{num decoded tokens} \cdot \text{FLOPs per token})\) represents the cost of generating and processing proposals in multiple aggregation layers, while \((\text{num decoded tokens} \cdot \text{FLOPs per token})\) corresponds to the final generation step.



\input{tex_sections/tables/case}

\end{document}

%% file: tex_sections/1-Introduction.tex
\section{Introduction}
\label{sec:intro}


Large language models (LLMs)~\cite{llama3,gpt4} have demonstrated remarkable success in applications ranging from coding~\citep{li2023starcoder,roziere2023code} and reasoning~\citep{yuan2023scaling,qin2024o1} to AI assistance~\citep{chiang2023vicuna}, driven by the expansion of both data and model capacity during pre-training~\citep{hoffmann2022training,scaling_law} and supervised fine-tuning (SFT)~\citep{yuan2023scaling,zhang2024when}. 
Beyond training-time scaling, recent studies~\citep{snell2024scaling,zhang2024scaling,moa} indicate that allocating additional compute at inference can further boost performance, enabling smaller models to rival larger ones. 
Correspondingly, a variety of inference-time scaling algorithms~\citep{madaan2023selfrefine,llm_monkey,swiechowski2023monte} have emerged, including sequentially revising generations or selecting among parallel samples.

\begin{figure}[t!]
\vskip 0.2in
\begin{center}
\centerline{\includegraphics[width=0.6\columnwidth]{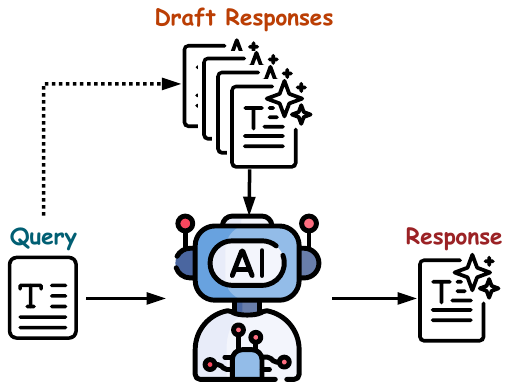}}
\caption{
Aggregation Fine-Tuning: Learning to generate a refined response by synthesizing multiple drafts alongside the query, different from the direct query-response mapping of traditional supervised fine-tuning. 
}
\label{fig:intro}
\end{center}
\vskip -0.2in
\end{figure}

Despite the growing complexity of inference algorithms, typical SFT pipelines still learn a direct mapping from a user query to a reference response. 
To this end, a line of work attempts to integrate the intermediate inference process such as rationale~\cite{star} and reasoning traces~\cite{tian2024toward,zhang2024restmcts}, into supervised training to improve test-time performance. 
While these intermediate steps have been shown effectiveness in reasoning tasks such as coding and math, they are less straightforward to extend to tasks such as writing or open-ended dialogue.
The intermediate process for general tasks remains less discussed. 

In this work, we seek to find an inference process for instruction-following and integrate it into supervised training to facilitate test-time scaling. 
Recently, ~\citet{moa} propose mixture-of-agents (MoA) which aggregates diverse response from multiple agents to improve generation quality.  
Drawing inspiration from the mixture-of-agents (MoA) framework~\citep{moa}, which aggregates diverse outputs from multiple LLMs, we propose to treat \emph{aggregation} itself as a general-purpose inference process for instruction following. 
This leads to a new supervised training framework, which we term \textbf{aggregation fine-tuning} (\textbf{AFT}).
As illustrated in Figure~\ref{fig:intro}, under AFT the model is trained to \emph{aggregate} multiple draft responses into a single, refined answer rather than directly mapping a query to a response. 
During inference, the model can itself generate and aggregate draft responses to produce a refined response.

To perform aggregation fine-tuning, we begin by constructing a specialized training dataset in which each instance comprises a query, multiple draft responses (i.e., \emph{proposals}), and a refined response (i.e., \emph{aggregation}). 
We consider two types of proposals: off-policy and on-policy proposals, differing on whether or not the proposals are sampled from the model to be trained. 
Subsequently, the model is trained to aggregate proposals to an aggregation given the query, thereby conventional fine-tuning of query-to-response alignment.
During inference, the model can perform \emph{propose-and-aggregate} in a multi-layer fashion, which can be regarded as MoA~\cite{moa} with only one agent. 
At each layer, the model aggregates its proposals at the previous layer and generate proposals for the next layer. 
This iterative process of proposal generation and aggregation allows the model to leverage both parallel sampling and sequential revision effectively.
Empirical results on Llama3.1-8B-Base and Mistral-7B-v0.1-Base demonstrate that AFT substantially surpasses standard SFT in various benchmark datasets.
On AlpacaEval 2, AFT models improve LC win rates by up to 13.9\% on average, 
with on-policy AFT models exhibiting particularly significant improvements.
Further performance gains are achieved through propose-and-aggregate at inference time, elevating Mistral-based models above a 30\% LC win rate and Llama-based models beyond 40\%.
Notably, the top-performing model, fine-tuned on only \textbf{64k} training examples \textbf{without preference alignment}, attains an LC win rate of \textbf{41.3\%}, surpassing strong competitors such as Llama3.1-405B-Instruct and GPT4.
In MT-Bench, propose-and-aggregate pushed the best-performing mode to a score of 8.1.
Additionally, AFT shows enhancements over SFT in downstream tasks such as GSM8K and IFEval.

Analytic findings reveal that by fitting answers conditioned on drafts, aggregation learning benefits from shaping a low-perplexity region, where the model converges in a faster and more stable way. 
We also observe that both the diversity and quality of proposals contribute to the aggregation quality. 
In addition, the propose-and-aggregate method effectively combines sequential refinement and parallel sampling to scale inference-time computation, producing consistent improvements across multiple tasks with manageable computational overhead. 
Overall, these insights highlight aggregation learning and propose-and-aggregate inference as a flexible, efficient strategy for unlocking the full potential of LLMs. 
We will release the code, data and model after the anonymous period.




%% file: tex_sections/2-RelatedWork.tex
\section{Related Work}
\paragraph{Supervised Fine-tuning of LLMs.}
Supervised fine-tuning plays a pivotal role in enhancing the instruction-following ability of LLM~\citep{alpaca_eval,mtbench,liu2024alignbench,ouyang2022training,qin2024infobench,jiang2024followbench,ifeval}. 
Previous works find that high-quality instruction-following data is essential for the success of SFT~\citep{zhao2024tree,liu2024what,xu2024wizardlm,zhou2023lima,li2024quantity} and various data-curation techniques are proposed~\citep{wang2023selfinstruct,du2023mods,chen2024alpagasus,li2024one,sun2023principledriven}. 
More specifically, including complex constrained~\citep{sun2024conifer,he2024complex}, multi-faced~\citep{lou2024muffin}, self-refined~\citep{star,cheng2024spar,wu2024meta} and sequential~\citep{hu2024fine} data could enhance the instruction-following ability of LLM. 
Additionally, several studies on reasoning suggest incorporating intermediate processes (e.g., refinements in sequential revision~\citep{cheng2024spar,wu2024meta} or search traces in step-level graph search~\citep{qin2024o1,xi2024enhancing,zhang2024restmcts}) into the SFT data to enhance performance in reasoning tasks. 
The proposed aggregation fine-tuning differs in that it uses aggregation as the inference process of general-purpose instruction-following tasks to improve test-time performance. 


\paragraph{Inference Scaling for LLMs.}
Recent research has shifted attention from training-time computation to test-time scaling. 
\citet{snell2024scaling} show that allocating more compute at inference can sometimes prove more effective than enlarging model size. 
Indeed, with existing inference-time algorithms~\citep{wei2022chain,yao2023tree,besta2024graph,chen2024universal,llm_monkey}, smaller models can surpass larger ones when both share a fixed test-time budget~\citep{wu2024empirical}. 
According to \citet{welleck2024decoding}, inference scaling can be broadly categorized into: 
\emph{sequential revision}, where an LLM functions as a callable module to iteratively refine its prior outputs~\citep{dohan2022language,khattab2024dspy,madaan2023selfrefine,shinn2023reflexion,welleck2023generating,havrilla2024glore}; 
\emph{parallel sampling}, which generates multiple candidate responses concurrently and selects the best via a ranker~\citep{llm_monkey,huang2024enhancing,jiang2023llm}; 
and \emph{step-level graph search}, wherein a complex query is decomposed into intermediate steps explored through tree-search~\citep{swiechowski2023monte} or graph-search algorithms~\citep{yao2023tree,besta2024graph,tian2024toward,chi2024thoughtsculpt}. 
While most of these methods focus on a single model's test-time scaling, \citet{moa} introduce a \emph{mixture-of-agents} approach that integrates multiple LLMs and aggregates their diverse outputs to boost performance. 
In this study, we bring MoA's aggregation concept into a single-model setting for both training and inference, where \emph{aggregation learning} uncovers an LLM's latent potential, and \emph{propose-and-aggregate} further elevates performance on par with models of substantially larger scale. 
In particular, we view aggregation learning as an intermediate process for general instruction following, while propose-and-aggregate unifies the strengths of sequential revision and parallel sampling at test time.

%% file: tex_sections/3-Method.tex
\begin{figure*}[t!]
\vskip 0.2in
\begin{center}
\centerline{\includegraphics[width=0.8\linewidth]{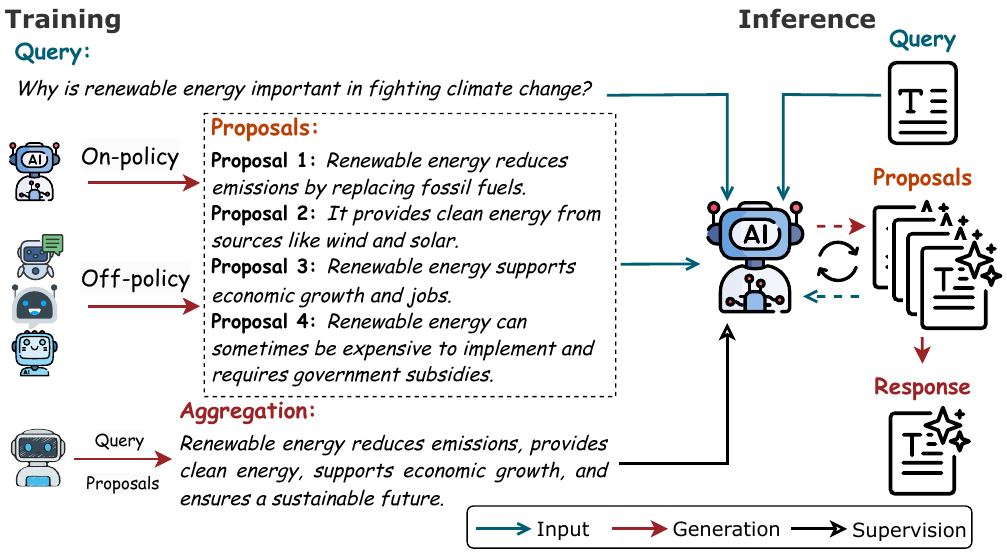}}
\caption{Framework of aggregation fine-tuning and propose-and-aggregate inference.}
\label{fig:method}
\end{center}
\vskip -0.2in
\end{figure*}

\section{Method}

In a traditional supervised fine-tuning setup, a language model $f_{\theta}: \mathcal{Q} \to \mathcal{R}$ is trained to map each query $q \in \mathcal{Q}$ directly to a single reference response $r^{*} \in \mathcal{R}$. 
In contrast, we propose a supervised training framework wherein a language model aggregates multiple draft responses, referred to as \textit{proposals}, into a refined final response, termed \textit{aggregation}. 
We define this process as \textbf{aggregation fine-tuning}.
Let $q \in \mathcal{Q}$ be a query, and $P = \{r_1, r_2, \dots, r_K\} \subseteq \mathcal{R}$ represent a set of $K$ proposals.
Given $(q, P)$, the goal of AFT is to train an aggregator $A_\theta: \mathcal{Q} \times \mathcal{R}^K \to \mathcal{R}$, parameterized by $\theta$, to produce a refined response $\hat{r} = A_\theta(q, P)$ that approximates a high-quality reference response $r^* \in \mathcal{R}$.
During inference, the model can \textit{iteratively} perform aggregation $A_\theta: \mathcal{Q} \times \mathcal{R}^K_{t-1} \to \mathcal{R}_t$ to refine responses, where $t$ is the iteration step or aggregation layer.

This process contrasts with traditional supervised fine-tuning, which directly maps a query $q$ to a single reference response $r^*$. 
By explicitly conditioning on the drafts $P$, the model requires a meta-skill which goes beyond producing an output for a query. 
It learns a reasoning process that evaluates, critiques, and improves upon initial solutions.
The overall framework is illustrated in Figure~\ref{fig:method}. 
We provide a detailed illustration of data construction, model training, and inference in the following sections. 


\subsection{Data Construction}
Constructing training data for aggregation fine-tuning involves collecting model proposals and a reference aggregation response, for each user query.

\paragraph{Proposal Collection.} 
We categorize proposals into two types: \textbf{off-policy} proposals and \textbf{on-policy} proposals.
\textit{Off-policy} proposals are derived from existing preference alignment datasets, such as UltraFeedback~\cite{cui2023ultrafeedback}, where each query is accompanied by multiple responses generated by models different from the one being fine-tuned.
\textit{On-policy} proposals are obtained by leveraging in-context learning (ICL) with demonstrations, prompting the base LLM to generate multiple responses for a given query~\cite{yejin_icl}. Since these proposals are sampled from the same LLM that will be fine-tuned, they are referred to as \textit{on-policy} proposals. 
In this way, the model learns how to revise and aggregate its own generations. 
Both methods result in a set of $K$ proposals, denoted as $P = \{r_1, \dots, r_K\}$, for each query $q$.

\paragraph{Aggregation Construction.} 
\label{sec:data_agg}
We use a dedicated prompt (Appendix~\ref{app:prompt}) to prompt a stronger language model to aggregate the proposals $P$ into a reference response $r^*$ that serves as the ground truth for training. 
This procedure ensures that $r^*$ demonstrates an ideal aggregation behavior, incorporating information from $P$ while correcting errors and improving quality. 
The data construction details can be found in Section~\ref{sec:exp}.

\subsection{Training}
The goal of aggregation fine-tuning is to maximize the probability that the model’s aggregator $A_\theta$ assigns to the reference aggregated response $r^{*}$. Formally, for a training dataset 
\(\{(q^{(i)}, P^{(i)}, r^{*(i)})\}_{i=1}^N\), where each example consists of a query \(q^{(i)}\), a set of proposals \(P^{(i)}\), and a high-quality reference aggregation \(r^{*(i)}\), we seek:
\begin{equation}
\max_\theta \; \frac{1}{N} \sum_{i=1}^N 
A_\theta\bigl(r^{*(i)} \!\mid\! q^{(i)},\, P^{(i)}\bigr).
\end{equation}
In contrast, conventional supervised training learns a mapping function $f$ from a query to reference response directly:
\begin{equation}
\max_\theta\frac{1}{N} \sum_{i=1}^N f_\theta(r^{*(i)} \mid q^{(i)})
\end{equation}

We use the same prompt as outlined in Section~\ref{sec:data_agg} to incorporate proposals.
Rather than producing an answer from scratch given only the query, the aggregator can build upon the semantic cues, linguistic structures, and partial solutions embedded in the drafts.
As a result, the model’s generation perplexity is significantly reduced, reflecting its ability to more efficiently predict the reference tokens due to the guidance offered by the existing candidates. 
Therefore, this enriched learning signal leads to a more stable and smooth training curve, as demonstrated in Section~\ref{sec:understand_agg}.



\subsection{Inference}
\label{sec:infer}
During inference, the AFT model can perform \textbf{propose-and-aggregate} to scale inference-time computing to boost performance.
The process of propose-and-aggregate involves iteratively aggregating proposals from the previous layer and generating aggregations, forming the proposals for the next layer. 

\paragraph{Initial Proposal Generation.} 
Given a query $q$, the model draws $K$ initial proposals $P^{(0)} = \{r^{(0)}_1, \dots, r^{(0)}_K\}$ by sampling:
\begin{equation}
    r^{(0)}_i \sim A_\theta(r|q) \quad i = 1,2,\ldots K.
\end{equation}
Choosing a single proposal from this set as the final response is equivalent to standard generation (i.e., $A_{\theta}(q)$) which does not scale inference-time computation.

\begin{figure}
    \centering
    \includegraphics[width=0.5\linewidth]{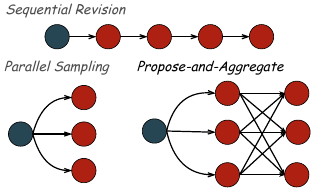}
    \caption{Test-time scaling of propose-and-aggregate compared with sequential revision and parallel sampling.}
    \label{fig:infer}
\end{figure}

\paragraph{Iterative Aggregation (Optional).}
Given the initial proposals, the model can optionally produce a refined response in a multi-layer setup. 
Specifically, the model can take in the aggregated responses from the previous layer $\ell$ (e.g., initial proposals at the first layer $P^{(0)}$) as input proposals:
\begin{equation}
P^{(\ell)} \;=\;\bigl\{\,r^{(\ell)}_1,\,\dots,\,r^{(\ell)}_K\bigr\} \quad \ell = 0,1,\dots,L,
\end{equation}
where $L$ is the maximum aggregation layer. 
Subsequently, the model performs aggregation on these proposals to further enhance generation quality:
\begin{equation}
r^{(\ell+1)}_i \;\sim\; A_\theta\bigl(r|q,\;P^{(\ell)}\bigr)  \quad i = 1,2,\ldots K,
\end{equation}
 forming the proposals for the next layer: $P^{(\ell+1)}$.
This iterative procedure continues until a predefined number of layers is reached.

Iterative propose-and-aggregate inference thus emulates a recursive reasoning process in which the model refines its responses over multiple stages. 
As noted by \citet{test_time}, LLMs can scale test-time computing via sequential revision or parallel sampling. 
As presented in Figure~\ref{fig:infer}, sequential revision sequentially refines model outputs to incrementally improve quality~\cite{revision}, whereas parallel sampling explores the sampling space under the guidance of a verifier~\cite{verify_step}.
Propose-and-aggregate combines the advantages of depth-first search (sequential revision) and breadth-first search (parallel sampling) by iteratively refining diverse aggregated generations from prior iterations. 
We demonstrate how our method scales along both depth and breadth axes in Section~\ref{sec:ana_scale}.




%% file: tex_sections/3.5-Experimental-Setup.tex
\section{Experimental Setup}
\label{sec:exp}
\paragraph{Data.}
We construct training data based on UltraFeedback~\cite{cui2023ultrafeedback}. 
UltraFeedback consists of 64k instances, each of which containing multiple model responses to a user query, providing a natural testbed for \textit{off-policy} proposals. 
Among these responses, we select the response with the highest score as the reference for training \textit{SFT} baselines.
To generate \textit{on-policy} proposals, we utilize in-context learning to prompt base LLMs (e.g., Llama-3.1-8B-base and Mistral-7B-v0.1-Base) for sampling 10 responses to a query in UltraFeedback, subsequently employing a reward model to select the top 5 highest-ranked samples.
We adopt the ICL template from prior research~\cite{yejin_icl}, and the reward model utilized is \texttt{sfairXC/FsfairX-LLaMA3-RM-v0.1}~\cite{rm}.
To promote generation diversity, we set the temperature to 1.0, top\_k to 50, and top\_p to 0.95, as recommended by~\citet{helpsteer}.
We use Qwen2.5-72B-Instruct~\cite{qwen2.5} to generate aggregated responses (aggregations) given queries and proposals following~\citet{moa}.

\paragraph{Training.}
We perform fine-tuning on two families of base LLMs: Llama-3.1-8B-Base~\cite{llama3} and Mistral-7B-v0.1-Base~\cite{mistral}. 
For both SFT and AFT models, training is conducted using Llama-Factory~\cite{zheng2024llamafactory}. 
All models are fine-tuned with LoRA~\cite{lora}, employing a rank of 8 and a total batch size of 128 across four A100 GPUs.
The learning rate is set to \(1 \times 10^{-4}\) with a cosine schedule featuring 10\% warmup steps for one epoch.

\paragraph{Evaluation.}
We primarily evaluate our models using two prominent open-ended instruction-following benchmarks: the single-turn dialogue evaluation AlpacaEval 2~\cite{alpaca_eval} and the multi-turn dialogue evaluation MT-Bench~\cite{mtbench}. 
These benchmarks assess the models' conversational capabilities across a diverse array of queries spanning various domains and tasks.
Scores are reported according to each benchmark's official evaluation protocol. 
For AlpacaEval 2, we present both the raw win rate (WR) and length-controlled win rate (LC)~\cite{lc}. 
For MT-Bench, we provide the average score with GPT-4 serving as the judging model.
Additionally, we also evaluate performance on downstream tasks, which encompass general knowledge (MMLU~\cite{mmlu}), reasoning abilities (ARC-c~\cite{arc} and StrategyQA~\cite{strategyqa}), mathematical skills (GSM8K~\cite{gsm8k}), and instruction following (IFEval~\cite{ifeval}). 
All evaluations adhere to evaluation configurations under OpenCompass~\cite{2023opencompass}, where we adpot few-shot generation configurations for MMLU, ARC-c, StrategyQA, and GSM8K, while zero-shot for IFEval. 
For IFEval, we report the average across all four scores.

\paragraph{Inference.}
For proposal generation and intermediate aggregation, we set the temperature to 0.7 and top\_p to 0.95 across all experiments. 
By default, we set the number of proposals to 5 and configure the aggregation layers to 2 for AlpacaEval 2 and MT-Bench, while employing a single-layer aggregation for GSM8K and IFEval.
For final response generation (i.e., last-layer aggregation), we configure settings according to standard routing practices.
For AlpacaEval 2, we utilize a temperature of 0.9 for Llama-based models and 0.7 for Mistral-based models following previous work~\cite{simpo}.
For MT-Bench, we adhere to the official decoding setup that prescribes varying sampling temperatures based on distinct categories.
For downstream tasks, we follow the OpenCompass configuration.

%% file: tex_sections/4-Experiments.tex
\section{Experimental Results}
In this section, we present the main results of our experiments, highlighting the enhanced performance achieved through aggregation learning and propose-and-aggregate inference-time scaling.

\input{tex_sections/tables/alpacal_eval}

\subsection{Main Results}

\paragraph{Performance on Single-turn Dialogues.}

The results on \textbf{AlpacaEval 2} are presented in Table~\ref{tab:alpaca}. 
For inference without scaling, both AFT-off-policy and AFT-on-policy demonstrate substantial improvements over their SFT counterparts, with an average LC win rate increase of 14.9\% for Mistral-based models and 12.9\% for Llama-based models.
For both model families, AFT-on-policy obtains stronger performance compared with AFT-off-policy. 
The performance advantage can be attributed to on-policy aggregation learning's resemblance to a process that refines the models' own drafts, positioning the model within a \emph{low-perplexity region} that aligns closely with its current capabilities and expectations (Section~\ref{sec:understand_agg}).
When scaling AFT models using the propose-and-aggregate method, the models achieve further performance improvements, with LC win rates exceeding 30\% for Mistral-based models and 40\% for Llama-based models.
The top-performing AFT model, AFT-on-policy trained from \textbf{Llama3.1-8B-Base}, achieves an LC win rate of \textbf{41.3\%}, surpassing several leading LLMs on the official leaderboard, including GPT-4 and \textbf{Llama3.1-405B-Instruct}.
Notably, our AFT models are fine-tuned solely on \textbf{64k} training instances \textit{without} further preference alignment using DPO~\cite{dpo} or RLHF~\cite{rlhf}, demonstrating how aggregation learning combined with propose-and-aggregate scaling elicit the potential of LLMs. 
We show in Section~\ref{sec:downstream} that SFT models without aggregation learning can hardly perform propose-and-aggregate to improve performance.
\input{tex_sections/tables/mt_bench}
\begin{figure}[t!]
\vskip 0.2in
\begin{center}
\centerline{\includegraphics[width=\columnwidth]{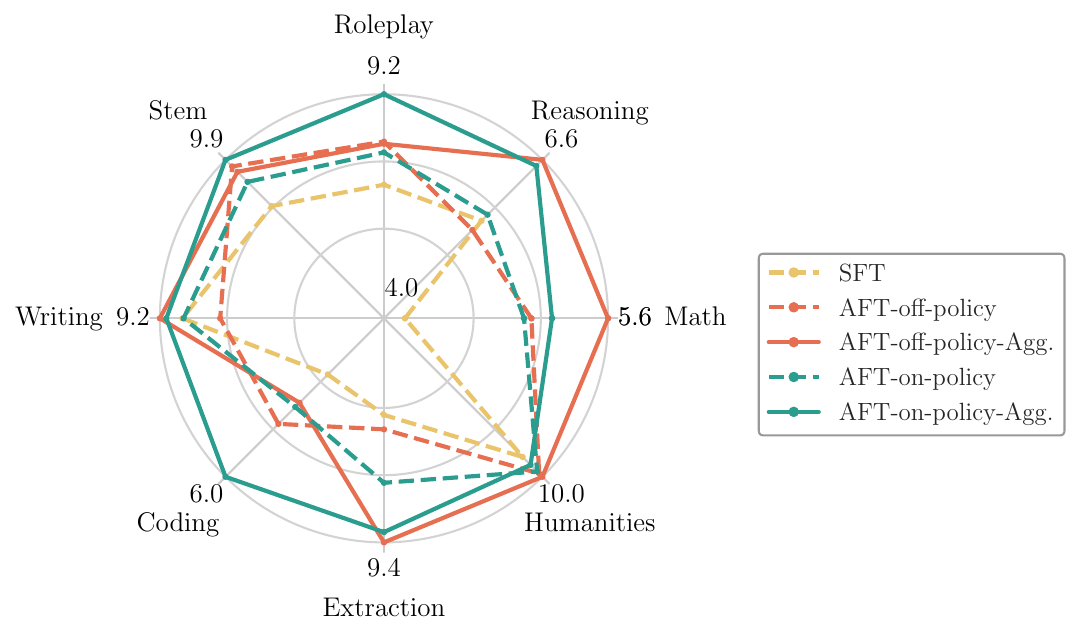}}
\caption{Radar chart for different models based on Llama3.1-8B-Base on MT-Bench.}
\label{fig:radar}
\end{center}
\vskip -0.2in
\end{figure}

\input{tex_sections/tables/downstream}

\begin{figure}[t]
    \centering
    \includegraphics[width=0.99\linewidth]{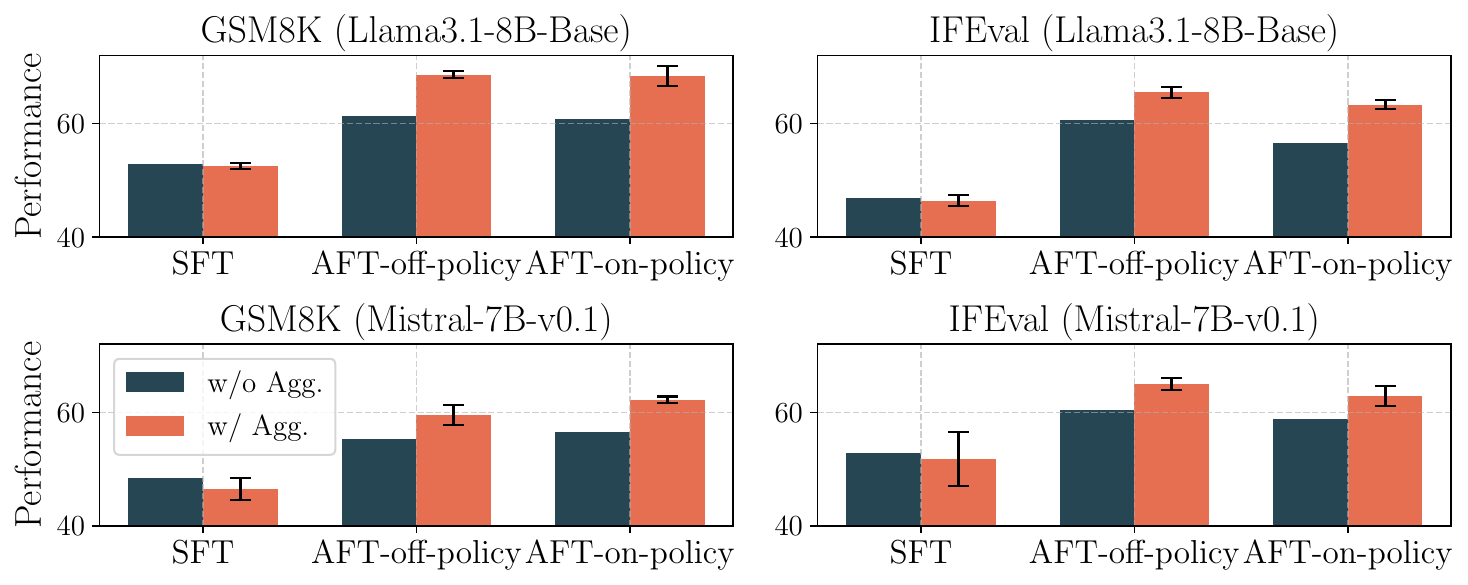}
    \caption{Performance of models fine-tuned on Llama-3.1-8B-Base on GSM8K and IFEval. ``w/ Agg.'' denotes inference using propose-and-aggregate.}
    \label{fig:gsm8k_ifeval}
\end{figure}

\paragraph{Performance on Multi-turn Dialogues.}
As shown in Table~\ref{tab:mtbench}, on \textbf{MT-Bench}, models utilizing aggregation learning consistently surpass the SFT baselines by a significant margin. 
For example, in Llama-based settings, AFT-on-policy attains an average score of 7.4, compared to 6.8 for SFT.
The propose-and-aggregate approach further enhances performance and elevates the top-performing model, AFT-on-policy (Llama3.1-8B-base), to an average score of \textbf{8.1}. 
The radar chart (Figure~\ref{fig:radar}) demonstrates that propose-and-aggregate consistently enhances instruction following ability across diverse tasks. 
We provide a case study in Appendix~\ref{app:case} to illustrate how propose-and-aggregate enhances generation quality.


\subsection{Downstream Task Performance}
\label{sec:downstream}
The evaluation results for downstream tasks are summarized in Table~\ref{tab:downstream} (MMLU, ARC-cA, and StrategyQA) and Figure~\ref{fig:gsm8k_ifeval} (GSM8K and IFEval).
AFT models demonstrate comparable performance to the SFT baseline on MMLU while consistently improving outcomes in reasoning tasks such as ARC-c and StrategyQA.
Due to the limited information provided in the answers for these tasks—specifically, single-choice options or yes/no responses—the propose-and-aggregate method is not applied. 
For GSM8K and IFEval, we perform propose-and-aggregate for three runs and report mean performance and standard errors. 
As shown in Figure~\ref{fig:gsm8k_ifeval}, AFT models consistently outperform SFT counterparts, and the advantages are further expanded with propose-and-aggregate. 
We can also observe that SFT models, without aggregation learning, \emph{cannot} perform propose-and-aggregate to boost performance. 

%% file: tex_sections/tables/alpacal_eval.tex

        
\begin{table}[t!]
\caption{Single-turn dialogue evaluation on \textbf{AlpacaEval 2}. $\dag$ denotes results obtained from the official leaderboard. ``w/ Agg.'' denotes inference using propose-and-aggregate.}
\label{tab:alpaca}
\vskip 0.15in
\begin{center}
\begin{small}
\begin{sc}        
        \begin{tabular}{lcc}  
            \toprule  
            \multirow{2}{*}{\textbf{Model}} & \multicolumn{2}{c}{\textbf{AlpacaEval 2}} \\
            & \textbf{LC(\%)}  & \textbf{WR(\%)}  \\
            
            \midrule
            Mistral Large (24/02)$\dag$  &	32.7 & 21.4 \\
            GPT-4 (03/14)$\dag$ & 35.3 & 22.1 \\
            Llama-3.1-70B-Instruct$\dag$ & 38.1 & 39.1 \\
            Qwen2-72B-Instruct$\dag$ & 38.1 &	29.9 \\
            GPT-4$\dag$ & 38.1 & 23.6 \\
            Llama 3.1-405B-Instruct &	39.3	& 39.1 \\
            Claude 3 Opus (02/29)$\dag$ & 40.5 &	29.1 \\
            \midrule
            \multicolumn{3}{c}{\textbf{Mistral-7B-v0.1-Base}} \\
            \midrule
            SFT   & 6.7 & 6.1  \\
            \noalign{\vskip 1.5pt}\hdashline\noalign{\vskip 1.5pt}
            AFT-off-policy & 19.8 &20.0 \\
             w/ Agg. &  \textbf{33.8}  &  47.8 \\
             \noalign{\vskip 1.5pt}\hdashline\noalign{\vskip 1.5pt}
            AFT-on-policy & 23.4 & 24.9 \\
            w/ Agg. &  30.7 & \textbf{48.4} \\
             \midrule
            \multicolumn{3}{c}{\textbf{Llama3.1-8B-Base}} \\
            \midrule
            SFT & 8.0 & 7.3 \\
            \noalign{\vskip 1.5pt}\hdashline\noalign{\vskip 1.5pt}
            AFT-off-policy & 20.3  & 19.6 \\
             w/ Agg. & 40.3 & 47.8 \\
             \noalign{\vskip 1.5pt}\hdashline\noalign{\vskip 1.5pt}
            AFT-on-policy & 21.5 & 21.8  \\
            w/ Agg. & \textbf{41.3} & \textbf{51.3} \\
    \bottomrule
    \end{tabular}
    \end{sc}
    \end{small}
    \end{center}
    \vskip -0.1in
    \end{table}

%% file: tex_sections/tables/mt_bench.tex

\begin{table}[t!]
\caption{Multi-turn dialogue evaluation on\textbf{ MT-Bench}. $\dag$ denotes results obtained from the official leaderboard.  ``w/ Agg.'' denotes inference using propose-and-aggregate.}
\label{tab:mtbench}
\vskip 0.15in
\begin{center}
\begin{small}
\begin{sc}        
        \setlength{\tabcolsep}{1.6pt} 
        \begin{tabular}{lccc}  
            \toprule  
             \multirow{2}{*}{\textbf{Model}} & \multicolumn{3}{c}{\textbf{MT-Bench}} \\
             & \textbf{1st turn}  & \textbf{2nd turn} & \textbf{Avg.} \\
            \midrule
            Mistral-7B-Instruct-v0.1$\dag$ & - & - & 6.8 \\
            GPT-3.5-Turbo-0314$\dag$ &- &- &7.9 \\
            Claude-2.0$\dag$ & - & - & 8.1 \\
            \midrule
            \multicolumn{4}{c}{\textbf{Mistral-7B-v0.1-Base}} \\
            \midrule
            SFT   & 6.6 & 6.1 & 6.4 \\
            \noalign{\vskip 1.5pt}\hdashline\noalign{\vskip 1.5pt}
            AFT-off-policy & 7.7 & 6.3 & 7.0\\
            w/ Agg. & 8.0 & 7.0 & 7.5\\
            \noalign{\vskip 1.5pt}\hdashline\noalign{\vskip 1.5pt}
            AFT-on-policy & 7.5 & 6.4 & 6.9\\
            w/ Agg. & \textbf{8.3} & \textbf{7.0} & \textbf{7.6} \\
            \midrule
            \multicolumn{4}{c}{\textbf{Llama3.1-8B-Base}} \\
            \midrule
            SFT & 7.3 & 6.2 & 6.8\\
            \noalign{\vskip 1.5pt}\hdashline\noalign{\vskip 1.5pt}
            AFT-off-policy & 7.7 & 6.9 &7.3 \\
             w/ Agg. & 8.3 & \textbf{7.6} & 7.9\\
             \noalign{\vskip 1.5pt}\hdashline\noalign{\vskip 1.5pt}
            AFT-on-policy & 7.9 & 6.9 & 7.4 \\
            w/ Agg. & \textbf{8.5}& \textbf{7.6}& \textbf{8.1}\\
\bottomrule
\end{tabular}
\end{sc}
\end{small}
\end{center}
\vskip -0.1in
\end{table}

%% file: tex_sections/tables/downstream.tex
        
\begin{table}[t]
\caption{Downstream task performance (MMLU, HellaSwag and ARC-c).}
\label{tab:downstream}
\vskip 0.15in
\begin{center}
\begin{small}
\begin{sc}     
        \setlength{\tabcolsep}{2pt} 
        \begin{tabular}{lccc}  
            \toprule  
            \textbf{Model} & \textbf{MMLU}   & \textbf{ARC-c} & \textbf{StrategyQA} \\
            \midrule
            \multicolumn{4}{c}{\textbf{Mistral-7B-v0.1-Base}} \\
            \midrule
            SFT   &  61.1 & 77.3 &  69.6 \\
            \noalign{\vskip 1.5pt}\hdashline\noalign{\vskip 1.5pt}
            AFT-off-policy & \textbf{61.6} & 80.0 & 70.1 \\
            AFT-on-policy & 61.1 & \textbf{80.7}  & \textbf{70.9} \\
            \midrule
            \multicolumn{4}{c}{\textbf{Llama3.1-8B-Base}} \\
            \midrule
            SFT & 62.3 & 78.6 & 64.1 \\
            \noalign{\vskip 1.5pt}\hdashline\noalign{\vskip 1.5pt}
            AFT-off-policy & \textbf{63.1} & \textbf{79.6} & 66.1  \\
            AFT-on-policy &  61.8 & 78.6 & \textbf{66.8}  \\
            \bottomrule
\end{tabular}
\end{sc}
\end{small}
\end{center}
\vskip -0.1in
\end{table}
     


%% file: tex_sections/5-Analysis.tex
\section{Analysis}
In this section, we delve into aggregation learning and the propose-and-aggregate framework.
We begin by examining how aggregation learning outperforms traditional supervised fine-tuning (Section~\ref{sec:understand_agg}).
Subsequently, we analyze key proposal patterns that impact aggregation quality (Section~\ref{sec:ana_tradeoff}), followed by experiments assessing test-time scaling in terms of search width and depth (Section~\ref{sec:ana_scale}).
Finally, we discuss the computational overhead of our method (Section~\ref{sec:ana_overhead}).
\subsection{Understanding Aggregation Learning}
\label{sec:understand_agg}

\begin{figure}[t!]
\vskip 0.2in
\begin{center}
\centerline{\includegraphics[width=0.99\columnwidth]{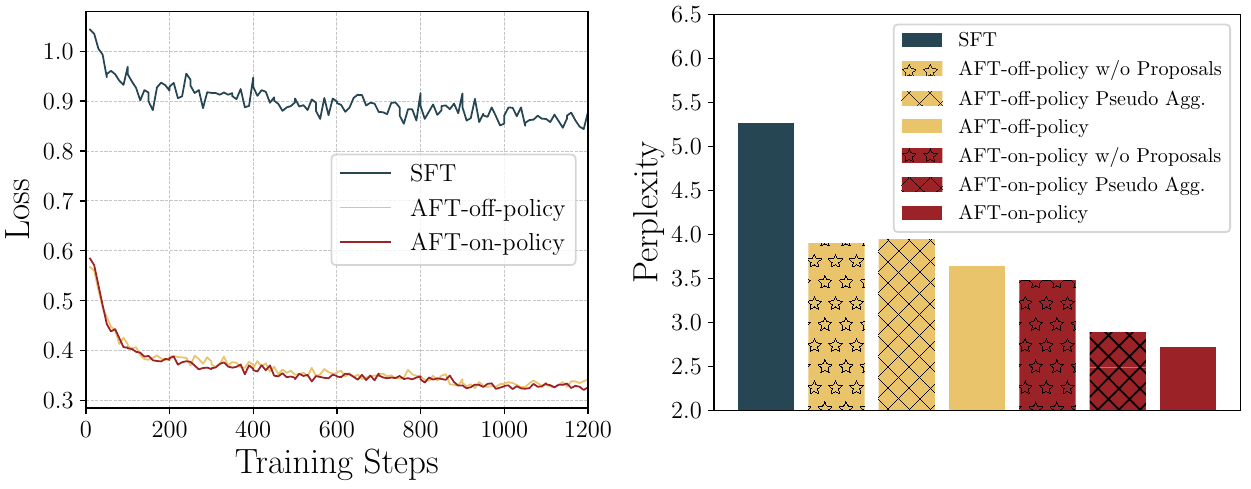}}
\caption{\textbf{Left}: training curves of SFT and AFT models (Llama3.1-8B-Base); \textbf{Right}: the perplexity of the base LLM of different training sets \emph{before} fine-tuning, where ``w/o Proposals'' indicates removing proposals when calculating perplexity, and ``Pseudo Agg.'' denotes using a pseudo aggregation in replacement of the real aggregation aggregated from proposals.}
\label{fig:train_loss_ppl}
\end{center}
\vskip -0.2in
\end{figure}

The training curves in the left portion of Figure~\ref{fig:train_loss_ppl} demonstrate that aggregation fine-tuning is more efficient and stable compared to standard SFT. 
AFT achieves lower training loss and converges faster while maintaining smoother progression with minimal fluctuations. These results suggest that AFT exerts smaller perturbation on the existing distribution of the base model. 

We sample 1,000 instances from different training dataset and calculate the perplexity of the base model before supervised training, as illustrated in the right section of Figure~\ref{fig:train_loss_ppl}.
Notably, perplexity on AFT data is significantly lower than that on SFT data, suggesting that aggregation learning resembles ``mode-seeking'', which results in a quicker accumulation of probability mass on a subset of high-reward responses during learning~\cite{why_los_ppl}. 
The underlying intuition is that predicting the final answer after reviewing draft responses (i.e., proposals) is considerably less uncertain compared to predictions made without such context. 
Consequently, AFT-on-policy further reduces perplexity since drafts are generated from the base model itself. 
This behavior corroborates observations from the training curves, further highlighting the advantages of fine-tuning of the aggregation.

In addition, we consider two \textbf{ablation} variants of AFT training: (1) predicting aggregation without proposals and (2) using direct responses from Qwen2.5-72B-Instruct as pseudo aggregations. 
The first variant resembles knowledge distillation from a stronger model, while the second disregards any underlying connection between proposals and aggregation.
As shown in Figure~\ref{fig:train_loss_ppl}, both removing intermediate proposals and employing pseudo aggregations shift the model away from its ``comfortable region'', characterized by higher perplexities.
Consequently, models trained under both variants experience performance degradation, as detailed in Table~\ref{tab:ablation}.



    


\begin{table}[t!]
\caption{Ablation study (Llama3.1-8B-Base): ``without Proposals'' indicates aggregation learning with proposals removed while ``Pseudo Aggregation'' denotes learning a pseudo ``aggregation'' target which is not aggregated from proposals.}
\label{tab:ablation}
\vskip 0.15in
\begin{center}
\begin{small}
\begin{sc}     
        \setlength{\tabcolsep}{2pt} 
        \begin{tabular}{lcc}  
            \toprule  
            \textbf{Model} & \textbf{GSM8K}  & \textbf{IFEval} \\
            \midrule
            AFT-on-policy & \textbf{60.7} &  \textbf{56.5} \\
            \noalign{\vskip 1.5pt}\hdashline\noalign{\vskip 1.5pt}
            without Proposals & 54.1 & 48.3 \\
            Pseudo Aggregation & 55.4 & 53.8 \\ 
            \bottomrule
\end{tabular}
\end{sc}
\end{small}
\end{center}
\vskip -0.1in
\end{table}

\subsection{Effects of Proposal Diversity and Quality}
\label{sec:ana_tradeoff}

\begin{figure}[t]
\vskip 0.2in
\begin{center}
\centerline{\includegraphics[width=0.7\columnwidth]{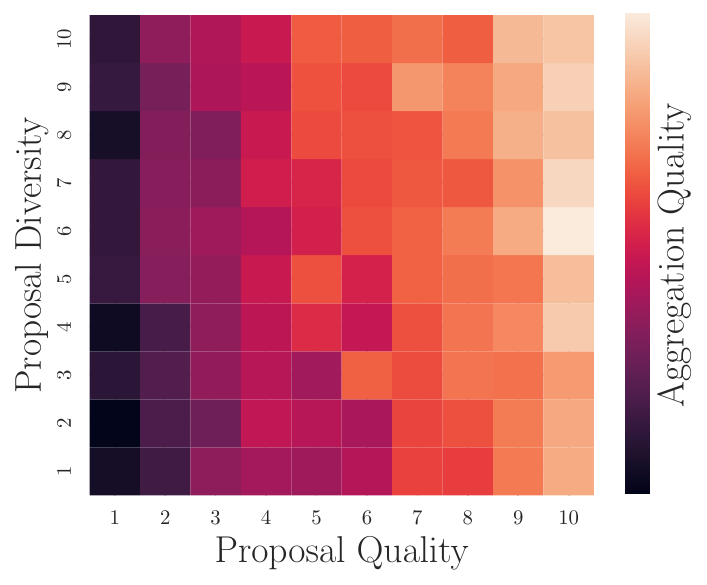}}
\caption{Aggregation quality in relation to proposal quality and diversity: each grid cell represents a score for both dimensions. A lighter color indicates higher aggregation quality.}
\label{fig:quality_div}
\end{center}
\vskip -0.2in
\end{figure}

We investigate the effects of proposal diversity and quality on aggregation quality. 
Using 100 instances sampled from AlpacaEval 2, we employ a reward model to evaluate the quality of generated responses (AFT-on-policy based on Llama3.1-8B-Base). 
For each query, we sample 10 proposals and systematically traverse all possible combinations of 5 proposals, resulting in altogether 25,200 groups of proposals. 
The model then generates aggregations for each combination, with the quality of these aggregations evaluated using the same reward model.
To measure proposal diversity, we utilize the Vendi score~\cite{vendi_score}, while the average quality of proposals within each combination serves as a measure of proposal quality. 
To enable comparisons across queries, we normalize the absolute scores by converting them into relative rankings on a scale of 1 to 10. 
The averaged results are presented in Figure~\ref{fig:quality_div}.

As shown in the figure, both the diversity and quality of proposals significantly impact aggregation performance. 
Proposals with higher diversity and quality generally lead to stronger aggregations. 
However, proposal quality plays a more dominant role, as evidenced by the more pronounced color variations along the horizontal axis. 
This indicates that higher-quality proposals contribute more substantially to aggregation quality. 

\subsection{Test-time Scaling along Width and Depth}
\label{sec:ana_scale}

\begin{figure}[t]
\vskip 0.2in
\begin{center}
\centerline{\includegraphics[width=\columnwidth]{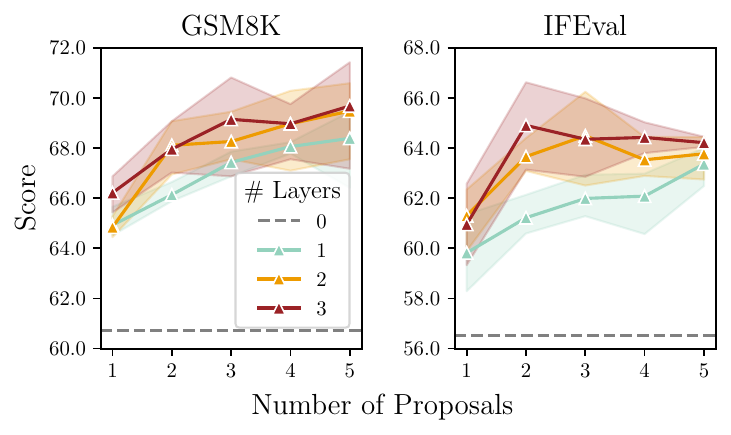}}
\caption{Performance w.r.t. aggregation width (number of proposals) and depth (number of aggregation layers).}
\label{fig:width_depth}
\end{center}
\vskip -0.2in
\end{figure}


As discussed in Section~\ref{sec:infer}, the propose-and-aggregate framework combines the strengths of sequential revision and parallel sampling to enhance inference performance. 
To analyze its scalability, we conduct experiments on AFT-on-policy (Llama3.1-8B-Base) using the GSM8K and IFEval datasets, averaging results across 3 runs. 
Figure~\ref{fig:width_depth} illustrates how test-time scaling affects performance, showing that increasing both the search width (number of proposals) and depth (number of aggregation layers) leads to performance improvements. 
Notably, the propose-and-aggregate framework degenerates into sequential revision when the number of proposals is set to one, and into parallel sampling when only a single aggregation layer is used. 
In the latter case, the model itself effectively acts as a verifier in the form of aggregation. 
These results highlight the flexibility of the propose-and-aggregate framework and its ability to adapt to varying computational and performance requirements.

\subsection{Computational Overhead} 
\label{sec:ana_overhead}


\begin{table}[t!]
\caption{Performance comparison between propose-and-aggregate and Best-of-N sampling (BoN) on models based on Llama-3.1-8B-Base. The aggregation layer $L$ and number of samples $N$ are set as 2 and 5 for propose-and-aggregate while BoN selects among 11 generations using a reward model (\texttt{FsfairX-LLaMA3-RM-v0.1}).}
\label{tab:rm}
\vskip 0.15in
\begin{center}
\begin{small}
\begin{sc}     
        \begin{tabular}{lccc}  
            \toprule  
            \multirow{2}{*}{\textbf{Model}} & \multicolumn{2}{c}{\textbf{AlpacaEval 2}} & \multirow{2}{*}{\textbf{FLOPs}} \\
            & \textbf{LC(\%)} & \textbf{WR(\%)} &  \\
            \midrule
            SFT & 8.0 & 7.3 & $F$\\
            w/ BoN (N=11) & 15.7 & 13.5 & $22F$\\
            \noalign{\vskip 1.5pt}\hdashline\noalign{\vskip 1.5pt}
            AFT-on-policy & 21.5 & 21.8 & $F$ \\
            w/ BoN (N=11)  & 32.7 & 35.4  & $22F$ \\
            w/ Agg. (L=2,N=5) & \textbf{41.3} & \textbf{51.3} & $21F$ \\
            \bottomrule
\end{tabular}
\end{sc}
\end{small}
\end{center}
\vskip -0.1in
\end{table}

We analyze the computational overhead of the propose-and-aggregate framework and compare with parallel sampling. 
We approximate inference FLOPs following previous work~\cite{llm_monkey} (Details in Appendix~\ref{app:flops}). 
The primary additional computational cost arises from the aggregation step, which processes all proposals from the previous layer as input prompts. 
The FLOPs for propose-and-aggregate, denoted as $\hat{F}$, can be approximated as:
    $\hat{F} \approx L \cdot (2 \cdot N \cdot F) + F$,
where $F$ represents the FLOPs for vanilla generation, $L$ is the number of aggregation layers and $N$ denotes the number of parallel proposals. 
A common parallel sampling baseline is Best-of-N (BoN), which uses an external reward model to rank multiple generated responses. 
Assuming the reward model is the same size as the policy model, the FLOPs for BoN can be approximated by $\bar{F} \;\approx\; 2 \,\cdot\, N \,\cdot\, F$.

Table~\ref{tab:rm} compares propose-and-aggregate to BoN under comparative FLOPs. 
Even when equipped with BoN, an SFT model cannot surpass the performance of an AFT model.
Furthermore, propose-and-aggregate offers greater improvements on AFT compared to BoN, indicating the benefit of expanding along both width (parallel proposals) and depth (iterative aggregation).
Notably, propose-and-aggregate does not rely on an external reward model to achieve test-time scaling, thereby simplifying deployment while retaining robust performance gains.

%% file: tex_sections/6-Conclusion.tex
\section{Conclusion}

In this work, we introduced \emph{aggregation fine-tuning} as a paradigm that teaches language models to aggregate multiple draft responses (i.e., proposals) into a refined final answer (i.e., aggregation). 
By conditioning on diverse proposals, AFT encourages higher-order reasoning and demonstrates consistent gains over standard supervised fine-tuning across both instruction-following benchmarks and downstream tasks. 
Furthermore, our \emph{propose-and-aggregate} method leverages iterative inference to further improve performance without additional training, combining the merits of both sequential refinement and parallel sampling. 
Analysis showed that AFT reduces perplexity, stabilizes convergence, and benefits from carefully balancing proposal quality and diversity, making it a flexible and cost-effective approach to unlocking latent model capabilities.

%% file: tex_sections/tables/case.tex
\begin{table*}[]
    \centering
    \small

    \begin{tabular}{p{0.1\textwidth}p{0.9\textwidth}}
        \toprule
        Query   & When Sophie watches her nephew, she gets out a variety of toys for him. The bag of building blocks has 31 blocks in it. The bin of stuffed animals has 8 stuffed animals inside. The tower of stacking rings has 9 multicolored rings on it. Sophie recently bought a tube of bouncy balls, bringing her total number of toys for her nephew up to 62. How many bouncy balls came in the tube? Let's think step by step. Answer: 
         \\
         \midrule
         Proposals 1 & Sophie has a total of 62 toys, which includes 31 building blocks, 8 stuffed animals, and 9 stacking rings. To find out how many bouncy balls were in the tube, we subtract the total number of toys except for the bouncy balls from the total number of toys: 62 - (31 + 8 + 9) = 62 - 48 = 14 \newline The answer is 14 bouncy balls. \\
        \midrule
           Proposals 2 & The total number of toys Sophie has is 31+8+9+12 bouncy balls = 60. If she has 62 toys after adding the bouncy balls, the number of bouncy balls is 62-60 = 2. The answer is 2. However, there seems to be a mistake in the problem. The total number of toys Sophie has is 31+8+9+12, which is 60, not 62. Therefore, there is an error in the problem, and we cannot determine the number of bouncy balls without more information. \\
        \midrule
           Proposals 3 & To determine the number of bouncy balls in the tube, we need to find the number of toys Sophie had before she bought the tube of bouncy balls, and then calculate the difference between 62 and that number. The total number of toys Sophie has for her nephew is 62. The bag of building blocks has 31 blocks, the bin of stuffed animals has 8 stuffed animals, and the tower of stacking rings has 9 rings. Adding these together gives us:  31 + 8 + 9 = 48. Therefore, the number of bouncy balls in the tube is:\[ 62 - 48 = 14 \] The answer is 14. \\
        \midrule
           Proposals 4 &

           To determine the number of bouncy balls in the tube, we need to find the total number of toys Sophie has and subtract the number of toys she already has from it. First, let's list the number of each type of toy: \newline- Building blocks: 31
           \newline
           - Stuffed animals: 8
           \newline
           - Stacking rings: 9
           \newline
           - Bouncy balls: 62
           \newline \newline
           Adding these together, we get:
           \newline
           31 + 8 + 9 + 62 = 110
           \newline
           \newline
           Now, we know the total number of toys is 110. We also know that Sophie already has 62 toys, so the number of bouncy balls in the tube is:
           \newline
           110 - 62 = 48
           \newline
           \newline
           The answer is 48 bouncy balls. 
        \\
        \midrule
           Proposals 5 & Let's start by finding the number of toys Sophie had before buying the tube of bouncy balls.\newline The bag of building blocks has 31 blocks.\newline
           The bin of stuffed animals has 8 stuffed animals.\newline
           The tower of stacking rings has 9 multicolored rings.\newline
           The total number of toys before buying the tube of bouncy balls is 31+8+9 = 48.\newline If Sophie now has 62 toys, the number of bouncy balls she bought is 62 - 48 = 14.\newline
           The answer is 14. \\
        \midrule
            Aggregation  & To determine the number of bouncy balls in the tube, we need to calculate the total number of toys before the bouncy balls and then subtract that from the total number of toys after the bouncy balls.\newline
            1. **List the number of toys before the bouncy balls:**\newline
            - Building blocks: 31\newline   - Stuffed animals: 8\newline  - Stacking rings: 9 \newline 
            2. **Calculate the total number of toys before the bouncy balls:**\newline  \[   31 + 8 + 9 = 48   \] 3. **Determine the total number of toys after the bouncy balls:**\newline   - Total toys after bouncy balls: 62\newline 
            4. **Calculate the number of bouncy balls:** \[   62 - 48 = 14   \] 
            The number of bouncy balls in the tube is 14. \\
        \bottomrule
    \end{tabular}
       \caption{Case study of propose-and-aggregate (math).}
    \label{tab:case_math}
\end{table*}

\begin{table*}[]
    \centering
    \small

    \begin{tabular}{p{0.1\textwidth}p{0.9\textwidth}}
        \toprule
        Query   & Imagine you are participating in a race with a group of people. If you have just overtaken the second person, what's your current position? Where is the person you just overtook?
         \\
         \midrule
         Proposals 1 & If you have just overtaken the second person in a race, you are now in the first position. The person you just overtook is now in the second position. \\
        \midrule
           Proposals 2 & If you have just overtaken the second person in a race, you are now in the second position. The person you just overtook is in the third position. \\
        \midrule
           Proposals 3 & In a race, overtaking the second person means you have just passed the person who was initially in second place. Therefore, your current position is third, and the person you just overtook is now in fourth place. \\
        \midrule
           Proposals 4 &
            If you have just overtaken the second person in a race, you are now in the first position, and the person you just overtook is now in the second position.
        \\
        \midrule
           Proposals 5 &If you have just overtaken the second person in a race, you are now in second place. The person you just overtook is in third place. \\
        \midrule
            Aggregation  &  If you have just overtaken the second person in a race, your current position is the second place. The person you just overtook will now be in third place. \newline \newline  To break it down:\newline - Initially, you were in third place.\newline - You overtook the second person, which means you are now in second place.\newline - The person you overtook is now in third place.\newline \newline So, the positions are:\newline 1. You (initially third, now second)\newline 2. The person you just overtook (initially second, now third) \\
        \bottomrule
    \end{tabular}
      \caption{Case study of propose-and-aggregate (reasoning).}
    \label{tab:case_reasonning}
\end{table*}

\begin{table*}[]
    \centering
    \small

    \begin{tabular}{p{0.1\textwidth}p{0.9\textwidth}}
        \toprule
        Query   &  Which word does not belong with the others?\newline tyre, steering wheel, car, engine
         \\
         \midrule
         Proposals 1 & The word that does not belong with the others is ``tyre." The other words—steering wheel, car, and engine—are all parts of a vehicle, whereas a tyre (or tire) is a component of the car. Tyres (tires) are the rubber parts that come into contact with the road and provide traction and support for the vehicle. \\
        \midrule
           Proposals 2 & The word that does not belong with the others is ``tyre." The other words are all parts of a car, but a tyre is a tire. The correct answer is ``tyre." \\
        \midrule
           Proposals 3 &
           The word that does not belong with the others is ``tyre." The other words—steering wheel, car, and engine—are all components of a vehicle. A tyre (or tire, as it's spelled in American English) is the rubber part of the vehicle that provides traction and support for the wheels. \\
        \midrule
           Proposals 4 &
            The word that does not belong with the others is ``steering wheel." The other words all describe parts of a car, but a steering wheel is a part of a car, whereas ``steering wheel" is a compound word that describes the function of the part, rather than the part itself. The correct answer is ``steering wheel".
        \\
        \midrule
           Proposals 5 &
           The word that does not belong with the others is ``steering wheel." The other three words—tyre, car, and engine—are all parts of a vehicle, but a steering wheel is an essential component of a car. \\
        \midrule
            Aggregation  &  The word that does not belong with the others is ``car." The other words—tyre (tire), steering wheel, and engine—are all components of a vehicle.. \newline\newline- **Tyre (Tire):** The rubber part of a vehicle that provides traction and support for the wheels.. \newline- **Steering Wheel:** The part of a car that the driver uses to control the direction of the vehicle.. \newline- **Engine:** The power source of a vehicle, typically an internal combustion engine, which converts fuel into mechanical energy to drive the vehicle.. \newline\newline A car, on the other hand, is a complete vehicle, not a specific part of a vehicle. Therefore, ``car" does not fit as well with the other three components.'\\
        \bottomrule
    \end{tabular}
    \caption{Case study of propose-and-aggregate (knolwdge).}
    \label{tab:case_knowledge}
\end{table*}

\begin{table*}[]
    \centering
    \small

    \begin{tabular}{p{0.1\textwidth}p{0.9\textwidth}}
        \toprule
        Query   &  Write an email to my boss telling him that I am quitting. The email must contain a title wrapped in double angular brackets, i.e. [[title]].\newline First repeat the request word for word without change, then give your answer (1. do not say any words or characters before repeating the request; 2. the request you need to repeat does not include this sentence)
         \\
         \midrule
         Proposals 1 & [[Request for Resignation]]\newline\newline I am writing to inform you that I am resigning from my position at [Company Name]. I have enjoyed my time here and appreciate the opportunities and experiences I have gained. I am grateful for the support and guidance you have provided me over the years. \newline\newline I am leaving to pursue a new opportunity that aligns better with my career goals and personal interests. I would like to thank you for the time and effort you have invested in me and wish you and the company continued success. \newline\newline Please let me know if there is any information or assistance you need from me during my transition. \newline\newline Thank you for your understanding.\newline\newline Sincerely,\newline\newline [Your Name]. \\
        \midrule
           Proposals 2 & [[Quitting]]\newline I am quitting.  \\
        \midrule
           Proposals 3 &
         [[Request for Resignation]] I am writing to inform you that I am resigning from my position as [your position] at [company name]. This decision has been a difficult one, but I have decided that it is in the best interest of my personal and professional growth to pursue new opportunities. I appreciate the support and mentorship I have received during my time here, and I am grateful for the experiences and skills I have gained. Please let me know how you would like to proceed with the transition process. Thank you for your understanding. \\
        \midrule
           Proposals 4 &
          [[quitting]] I am writing to inform you that I am quitting my position as your assistant.
        \\
        \midrule
           Proposals 5 &
          [[Title]] I am resigning from my position as a Software Engineer effective immediately. I would like to thank you for the opportunities and support you have provided me over the past year, and I wish the company continued success in the future. \\
        \midrule
            Aggregation  &  [[Request for Resignation]]\newline\newline I am writing to inform you that I am resigning from my position as [your position] at [company name]. I have enjoyed my time here and appreciate the opportunities and experiences I have gained. I am grateful for the support and guidance you have provided me over the years. \newline\newline I am leaving to pursue a new opportunity that aligns better with my career goals and personal interests. I would like to thank you for the time and effort you have invested in me and wish you and the company continued success. \newline\newline Please let me know if there is any information or assistance you need from me during my transition. \newline\newline Thank you for your understanding.\newline\newline Sincerely,\newline\newline [Your Name]. \\
        \bottomrule
    \end{tabular}
     \caption{Case study of propose-and-aggregate (writing).}
    \label{tab:case_writing}
\end{table*}

\begin{table*}[]
    \centering
    \small

    \begin{tabular}{p{0.1\textwidth}p{0.9\textwidth}}
        \toprule
        Query   &  Embrace the role of Sheldon from ``The Big Bang Theory" as we delve into our conversation. Don’t start with phrases like ``As Sheldon". Let's kick things off with the following question: ``What is your opinion on hand dryers?"
         \\
         \midrule
         Proposals 1 & Sheldon Cooper: Well, I find hand dryers to be an intriguing concept, but I must admit that I prefer the hygienic and efficient nature of paper towels. Hand dryers can be quite noisy and often fail to achieve a thorough drying, leading to damp hands. Additionally, they can spread germs if not properly maintained. Paper towels, on the other hand, are quick, effective, and ensure that hands are completely dry and germ-free. Therefore, I maintain my preference for paper towels in this context. \\
        \midrule
           Proposals 2 & I find hand dryers to be highly inefficient and often unsanitary. They can spread germs and bacteria, and the noise they emit can be quite irritating. I prefer to use paper towels, which are more hygienic and quieter. What are your thoughts on this matter?  \\
        \midrule
           Proposals 3 &
        Sheldon Cooper: \newline\newline It's a valid question, and one that I've given considerable thought to. Hand dryers are a common fixture in public restrooms, but their efficiency and hygienic standards leave much to be desired. I prefer paper towels for a number of reasons:\newline
        1. **Hygiene**: Hand dryers can be a breeding ground for bacteria and germs. The warm air can actually spread germs and bacteria more easily than paper towels, which are generally more hygienic and more effective at drying hands quickly and thoroughly.\newline\newline
        2. **Noise**: Many hand dryers are loud and disruptive, which can be a significant inconvenience, especially in quiet settings like libraries or offices.\newline\newline
        3. **Effectiveness**: While hand dryers can dry hands, they often take longer and are not as thorough as paper towels, which can dry hands in a matter of seconds.\newline\newline
        4. **Environmental Impact**: Paper towels, when disposed of correctly, are biodegradable and less harmful to the environment compared to the energy consumption and noise pollution of hand dryers.\newline\newline
        5. **User Experience**: Paper towels are more user-friendly, especially for those with disabilities or who have trouble operating electronic devices.\newline\newline
        In summary, while hand dryers are a popular choice, they do not offer the same level of hygiene, quietness, efficiency, and environmental friendliness as paper towels. Therefore, I strongly recommend opting for paper towels in public restrooms. \\
        \midrule
            ...
        \\
        \midrule
            Aggregation  & Certainly! Let's dive into the discussion on hand dryers with Sheldon's perspective.\newline
            Sheldon Cooper: ``What is your opinion on hand dryers?"\newline
            Well, hand dryers are a common feature in public restrooms, and they certainly have their advantages and disadvantages. Here are my thoughts:\newline\newline
            1. **Hygiene**: One of the most significant concerns with hand dryers is their potential to spread germs. The warm air can blow bacteria and viruses into the air, which can then be inhaled or transferred to other surfaces. This can be a particular issue in public restrooms where hygiene standards are often not as stringent. In contrast, paper towels are generally more hygienic because they can be replaced frequently and are more effective at removing bacteria from the hands.\newline\newline
            2. **Noise**: Hand dryers can be quite noisy, which can be a significant annoyance, especially in quiet environments like libraries or offices. This noise can be disruptive and can be a source of frustration for those around you.\newline\newline
            3. **Effectiveness**: While hand dryers can dry hands, they often take longer and may not dry hands as thoroughly as paper towels. Paper towels are quick and efficient, providing a more thorough and reliable drying experience.\newline\newline
            4. **Environmental Impact**: Hand dryers are energy-efficient and can help reduce the environmental impact compared to paper towels. However, the energy consumption and the noise pollution they generate should also be considered. Paper towels, when disposed of correctly, are biodegradable and less harmful to the environment.\newline\newline
            5. **User Experience**: Paper towels are more user-friendly, especially for those with disabilities or who have difficulty operating electronic devices. They are easy to use and can be more intuitive for a wide range of users.\newline\newline
            In summary, while hand dryers have their advantages in terms of energy efficiency and cost-effectiveness, they do not offer the same level of hygiene, quietness, efficiency, and user-friendliness as paper towels. Therefore, I strongly recommend opting for paper towels in public restrooms for a more hygienic, quieter, and user-friendly experience. \\
        \bottomrule
    \end{tabular}
   \caption{Case study of propose-and-aggregate (role-play).}
    \label{tab:case_role_play}
\end{table*}